\documentclass[letterpaper, 10 pt, conference]{ieeeconf}  

\IEEEoverridecommandlockouts                              

\usepackage{graphics} 
\usepackage{graphicx}
\usepackage{soul,color}
\usepackage{amsmath}
\soulregister\cite7
\soulregister\ref7
\soulregister\pageref7
\usepackage{booktabs}
\usepackage{multirow}
\title{\LARGE \bf
	Temporal Point Cloud Completion with Pose Disturbance
}

\author{Jieqi Shi$^{1}$, Lingyun Xu$^{2}$, Peiliang Li$^{3}$, Xiaozhi Chen$^{4}$ and Shaojie Shen$^{5}$%
\thanks{$^{1}$Jieqi and $^{4}$Shaojie Shen are with Department of Electronic and Computer
Engineering, the Hong Kong University of Science and Technology
        {\tt\footnotesize $\{$jshias, eeshaojie$\}$@connect.ust.hk} $^{2}$Lingyun Xu, $^{3}$Peiliang Li and $^{4}$Xiaozhi Chen are with Dji Co.
        {\tt\footnotesize $\{$judy.xu, peiliang.li$\}$@dji.com, cxz.thu@gmail.com}. Digital Object Identifier (DOI): see top of this page.}
}
\begin{document}
	
	\maketitle
	\thispagestyle{empty}
	\pagestyle{empty}

	\begin{abstract}
	 Point clouds collected by real-world sensors are always unaligned and sparse, which makes it hard to reconstruct the complete shape of object from a single frame of data. In this work, we manage to provide complete point clouds from sparse input with pose disturbance by limited translation and rotation. We also use temporal information to enhance the completion model, refining the output with a sequence of inputs. With the help of gated recovery units(GRU) and attention mechanisms as temporal units, we propose a point cloud completion framework that accepts a sequence of unaligned and sparse inputs, and outputs consistent and aligned point clouds. Our network performs in an online manner and presents a refined point cloud for each frame, which enables it to be integrated into any SLAM or reconstruction pipeline. As far as we know, our framework is the first to utilize temporal information and ensure temporal consistency with limited transformation. Through experiments in ShapeNet and KITTI, we prove that our framework is effective in both synthetic and real-world datasets. 
		
	\end{abstract}

	\section{INTRODUCTION}
    The 3D point cloud completion task aims to output the complete shape of objects from partial 3D inputs, and is of great significance to many subsequent tasks, such as robot navigation, AR reconstruction, and pose estimation. Previous researchers have examined reconstructing a complete object from single-frame point clouds, and have developed methods for different input data, such as voxels\cite{Dai2017ShapeCU,Han2017HighResolutionSC,Xie2020GRNetGR}, point clouds\cite{Yuan2018PCNPC,Wen2020PMPNetPC} and implicit fields\cite{Gao2020LearningDT,Deng2020DeformedIF}. However, previous methods have limitations that prevent them from being widely used in real-world tasks. 
    
    One limitation of the existing methods is that they assume that the input data are aligned to the canonical frame, where objects all face forward and are centered at the origin of the coordinates. This assumption is strict and is hard to obey in self-collected datasets, such as self-driving driving datasets collected by common lidar sensors. A second limitation is that researchers assume that partial inputs from a single frame are sufficient for accurate shape completion. In simulation datasets, e.g., ShapeNet\cite{Chang2015ShapeNetAI}, partial point clouds are collected by projecting a CAD model into certain angles for dense depth images\cite{Yuan2018PCNPC}, which ensures that the partial input contains enough geometric information of objects. However, real-world data are often sparse and randomly sampled, and the point clouds from a single frame may be meaningless. Last but not least, researchers often do not consider the consistency between the complete shapes recovered from different partial inputs of the same object, which is fatal to the scene reconstruction task. 
	\begin{figure}[t]
		\centering
		\framebox{\parbox{3.2in}{
				\centering

				\includegraphics[scale=0.255]{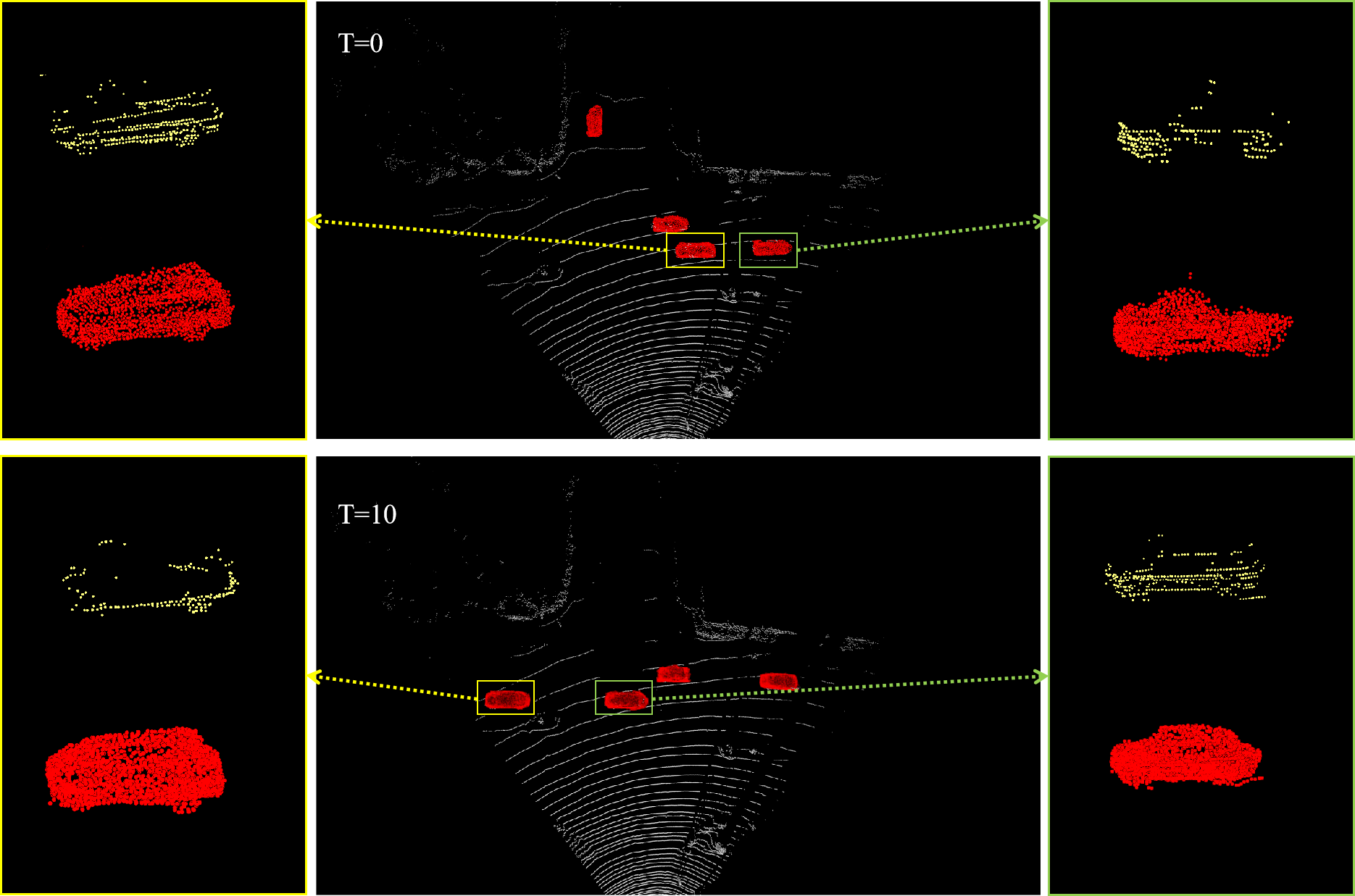}
		}}
		\caption{Illustration of our temporal completion results. Upper: Completion results of one randomly selected frame in KITTI test seq. 17. Lower: Completion results after 10 frames. Bounding boxes are from AB3DMOT.}
		\label{panda}
	\end{figure}	
	
	From our point of view, the last two problems come from the inadequate use of temporal information. In actual scenes, we can usually get enough frames of data for initialization, though each frame may have errors and shortcomings. Considering this, how to use temporal information effectively is of great significance for real-world tasks. Previously, \cite{Gu2020Weaklysupervised3S}  employed subsequent frames to train real-world models in a weakly-supervised manner and has provided promising result. However, they still focus on single-frame models and can not utilize multi-frame inputs to enrich the reconstructed model. In this paper, we want to delve deeper, utilizing the temporal information of multi-frame point clouds while tolerating certain disturbance in poses. 
	
	To better use the temporal information and handle the pose disturbance, we propose two main modifications to existing completion frameworks. 1) Design a backbone that is capable of dealing with pose disturbance of unaligned input data. In most real-world pipelines, automatic 3D detection methods, such as deep learning methods e.g. PVRCNN\cite{PVRCNN}, are employed to crop out interested objects. However, even the state-of-the-art(SOTA) detection methods still suffer from rotation and translation bias, which is critical for 3D completion tasks. Targeting this problem, we try to transform the unaligned point cloud to uniform coordinates while outputting a coarse completion result. 2) Propose several temporal units to fully fuse partial input from different frames and output a consistent result. As mentioned above, the input point clouds collected by real-world lidar sensors are often sparse and randomly sampled, which means that point clouds of subsequent frames are completely different and have few overlaps. One natural way to solve this problem is to accumulate points together after alignment. This idea is employed by \cite{Pang2021ModelfreeVT}, where the authors calculate the quality of incoming point clouds and overlay them over the first frame. However, experimental results show this method heavily relies on precision of the tracking pipeline which leads to considerable cumulative error. Also, much memory is required to recollect the point cloud information. To overcome this, we employ the popular memory gate strategy to memorize the geometric information and fuse them at the feature level. In order to make better use of recent data, we further added sliding window to enhance short-term memory in the refinement stage. We also want to complete the point clouds in an on-line manner, which means that we will output a complete result upon the arrival of each frame, and optimize the output results as the number of input frames increases.

	In this paper, we propose a temporally consistent 3D point cloud completion model that accepts unaligned point cloud sequences as inputs. We first employ an encoder-decoder framework for rough completion and pose recovery. Then we refine the initial shape with aligned partial point clouds. Temporal units are combined in both steps to maintain the consistency of the generated shape between frames. To demonstrate the effectiveness of our method, we conduct experiments on both the synthetic ShapeNet datset\cite{Chang2015ShapeNetAI} and real-world KITTI dataset\cite{Geiger2012AreWR,Geiger2013VisionMR}, which proves that our model is simple but effective. Our key contributions are:
	\begin{itemize}
		\item[1.] Propose a simple but efficient network for the unaligned point cloud completion task and obtain the SOTA  quantitative results. The pipeline can be integrated to any detection pipelines without further alignment, for automatic shape generation.
		\item[2.] Fully utilize temporal information to provide robust complete point clouds, with limited memory usage.
		\item[3.] Develop our framework to accept a arbitrary number of input frames and operate in an on-line manner. It processes each frame of a point cloud and provides a refined result upon arrival. Only a small number of inputs are needed to present a satisfactory result, which vastly speeds up the system initialization.
	\end{itemize}

    \begin{figure*}[ht]
		\centering
		\vspace{1em}
		\framebox{\parbox{6.6in}{
				\includegraphics[scale=0.56]{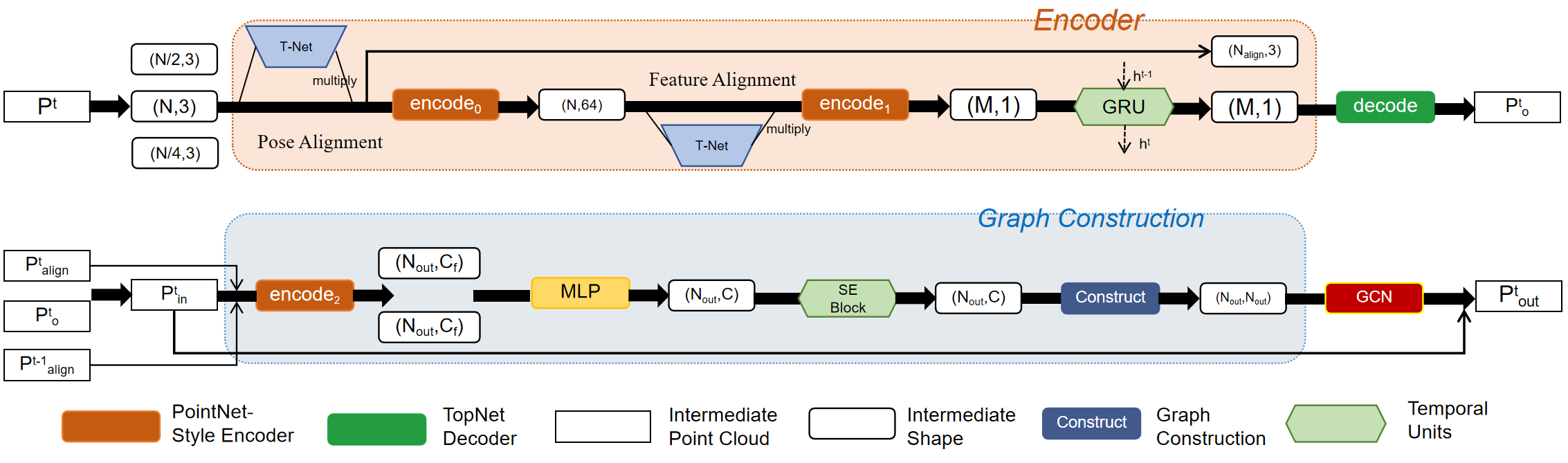}
		}}
		\caption{Details of our framework. The main modifications are highlighted in different colors. Upper: Alignment and completion. Lower: Shape refinement.}
		\label{framework}
	\vspace{-2em}
	\end{figure*}

	\section{Related Work}
	Though targeting the 3D point cloud completion task, our method is related to some other 3D vision tasks. In this section, we give a brief overview of related topics and differences between our work and previous methods.
    
	\subsection{Structure from Motion and Multi-View Stereo}
	The traditional structure-from-motion(SFM) task is often part of the multi-view stereo(MVS) reconstruction in terms of their pipelines. Both tasks accept an image sequence as input and try to recover the depth. SFM tasks usually pay more attention to robust key-points, and utilize feature matching methods to get 3D landmarks. Such sparse landmarks can be further used to optimize the pose of sensors in bundle adjustment\cite{Snavely2006PhotoT,Snavely2007ModelingTW} and SLAM\cite{MurArtal2015ORBSLAMAV} problems. The traditional SFM tasks have been further extended to deep-VO\cite{Li2020SelfSupervisedDV,Xue2019BeyondTS}, which extracts the poses of sensors from a sequence of images. Meanwhile, MVS problems go a step further to recover dense depth. Traditional MVS methods can be roughly divided into region-growing methods and depth-fusion methods. The region-growing methods, such as PMVS\cite{Furukawa2007AccurateDA}, rely on the result of robust key-points to generate patches and later extend to nearby regions. The depth-fusion methods, such as Kinect Fusion\cite{Newcombe2011KinectFusionRD}, on the other hand, process each frame as a whole and fuse the information of different frames at the depth level. This strategy is inherited by popular deep-learning methods. For example, MVSNet\cite{Yao2018MVSNetDI} constructs a dense cost-volume for the reference image using the features extracted by the network and calculates the depth of a whole image, and DeepMVS\cite{Huang2018DeepMVSLM} generates a group of plane-sweep volumes and predicts the disparity between input images. 
	
	Our task is similar to the SFM and MVS problems in that it also tries to fuse the 3D information of different frames and provide a refined shape. However, different from these traditional 3D problems, which take dense 2D images as input, we utilize sparse 3D point clouds, which brings about different challenges. Most importantly, it is hard to find key-points in 3D point clouds and operate 3D registration before fusion with randomly sampled inputs that may have scarcely any overlap. We, therefore, can not follow the popular extract-and-match pipeline that depends on landmarks, and must find another method for temporal fusion.
	
    \subsection{3D Shape Completion}
    
	The 3D shape completion task is attracting more and more researchers, who have been making headway on this problem. The research focuses on different input data types, such as point clouds\cite{Wen2020PMPNetPC,Yin2018P2PNET,Pan2021VariationalRP}, voxels\cite{Dai2017ShapeCU}--\cite{Xie2020GRNetGR,Stutz2018Learning3S}, and SDF\cite{Gao2020LearningDT,Thai20203DRO,Liu2020DISTRD} , and popular methods follow a standard encoder-decoder framework to get a rough result, often adding auxiliary tasks to enhance the shape code generation process\cite{Dai2017ShapeCU, Park2019DeepSDFLC}. To refine the generated point clouds, researchers have made efforts in optimizing the training methods\cite{Xia2021ASFMNetAS,Wen2021Cycle4CompletionUP}, utilizing more strict loss functions\cite{Xie2021StylebasedPG,jieqigcc}, adding information from another sensor for more detail\cite{Zhang2021ViewGuidedPC}, and developing better encoder networks\cite{Wen2020PMPNetPC,Pan2021VariationalRP}. However, most of these methods are built upon the assumption that the input data are well-aligned and each frame contains enough information for shape reconstruction. While this assumption can be easily satisfied in synthetic datasets, it is nearly impossible to meet in real-world problems. 
	
	Recognizing the limitations of popular 3D shape completion methods, \cite{Gu2020Weaklysupervised3S} utilizes the temporal information of more frames. However, different from our method, which enriches the shape information using several frames, the authors utilize subsequent frames to train the model in a weakly-supervised manner, taking the distance between the shape reconstructed from one frame and the input from another frame as the guidance signal. Yet the idea is inspiring and proves that, though not as consistent as 2D images, 3D point clouds still maintain consistency between frames. Moreover, \cite{Pang2021ModelfreeVT} integrates the shape completion task into the 3D tracking pipeline, and uses accumulated point clouds as the constructed shape. Such a fusion strategy is simple but noisy, leading to cumulative error similar to that in a SLAM system. The author also points out that memorizing past point cloud information consumes too many memory resources. To solve the problem, the author selects to use key frames instead of memorizing all input data, yet the memory usage is still troublesome and should be improved.

	\section{Method}
	 Most point cloud completion framework assumes the input points are collected in a unified reference frame. However, in actual case, the temporal object point clouds can not be well aligned due to the pose estimation error. Our method is designed to take temporal object point clouds  as input, where the point cloud of each frame has a non-trivial rotation error with less than 20 degree, and translation error with less than 0.1m with respecting to the reference frame, and output aligned and completed object point cloud. Following the existing GCC\cite{jieqigcc}, our network can be divided into an alignment and completion network, and a refinement network, and modify each stage by adding temporal units to memorize the temporal information. Therefore, our network can be treated as the combination of a per-frame processing backbone and the inserted temporal units, as illustrated in Fig \ref{framework}. In this section, we first introduce the per-frame processing units, and then explain in detail our temporal units and how they are used to encode the information of subsequent frames. We denote the input point cloud as $P$ with number of points $N$, rotation $e_r$, and translation $e_t$. We use footmarks to distinguish between the outputs of different stages of network. The output point cloud of the alignment network is denoted as $P_{align}$, and the rough completed point cloud of the stage-1 network is $P_{o}$. Similar as GCC, the input of the stage-2 network is a combination of the input point cloud and the output of the stage-1 network, and we name it $P_{in}$.

	\subsection{Per-Frame Completion}
	
	We also suppose the point clouds are scaled within a ball with radius = 0.5m, which follows the assumption of PCN\cite{Yuan2018PCNPC}. Considering the results from popular detection networks, such as PointRCNN and PVNet, we assume that the input point cloud $P$ has a rotation of $-20° < e_r < 20°$ and a translation of $-0.1m < e_t < 0.1m$. The basic framework of the stage-1 alignment and completion network is the same as that of GCC\cite{jieqigcc}. We extract the general shape features of the point cloud using a stacked Multilayer Perceptron(MLP) network and pooling layers of different resolutions. Different from the original network, we follow PointNet\cite{Qi2017PointNetDH} and add modified T-Net alignment units to the basic encoder. 
    
    The first unit is inserted at the beginning of the network. The original T-Net unit gets a $3 \times 3$ transformation matrix from a series of convolution-batchnorm-ReLU layers. However, such a design can not ensure the orthogonality of the matrix, which means that the rotation matrix may be meaningless and may even lead to unexpected deformation. Therefore, we make a slight modification and output 6D rotation and the 3D translation vector $T = (x, y, z)$ instead, following \cite{Gu2020Weaklysupervised3S}. After the network, we transform the 3D rotation vector into simple rotation matrix and do the alignment directly on $P$ to get aligned point clouds $P_{align}$, which are used as the final input of the encoder-decoder network. The second alignment unit is applied to features only. After several MLP layers in the ShapeNet encoder, we take out the intermediate features $f_{inter}$ of size $N \times C$ and feed them into another T-Net network. This time we output a transformation matrix of size $C \times C$, which is later multiplied to $f_{inter}$. The output of the encoder should be the aligned point clouds $P_{align}$, the translation $T$, and the shape code $S$ of dimension 1280. To extract the information from different resolutions, we apply furthest point sampling to $P$ and obtain three point clouds of size $N, \frac{N}{2}, \frac{N}{4}$, which means we will have three $P_{align, i}; T^{i}; i = 0, 1, 2$, and $S_{i}; i=0,1,2$. The shape code will be concatenated for a final $S_{final}$, and the channel will then be reduced by simple MLP layers. After the encoder network, we use the same decoder module as GCC\cite{jieqigcc}, which takes the shape code as input and outputs a rough complete point cloud $P_o$. The final output of the stage-1 network should be $P_{align, i}; T_i; i=0,1,2$, and $P_r$.
    
    As we have mentioned the three alighed point clouds of the stage-1 network are of different resolutions of size $N, \frac{N}{2}, \frac{N}{4}$, and the last two are too sparse and lose much geometric information. We therefore only feed the one of size $N$ into the stage-2 network. We then apply fps to $P_{align}^0$ and $P_r$ separately, form a new input point cloud $P_{in}$ of size $N_{out}$ with the selected points, and calculate the adjacent matrix $A$ of $P_{s2}$. We also extract points from $P_{align}$ that are within a ball centered at each point in $P_{in}$ and form a feature $f_{align}$ of size $N_{out} \times C_f$. This is the same as the procedure to combine controlling points with points to be optimized in \cite{jieqigcc}. These feature will then go through a graph convolutional network(GCN) for deformation. The only difference from the original deformation network is that we let the network regress a deformation between the input and target, and the final result should be the output of the GCN plus the input $P_{in}$.

	\subsection{Temporal Units}
	We use two kinds of units to fuse the temporal information between frames. The first is the global fusion operated in the alignmenet network, and the second is the local fusion in the refinement stage.
	
	Recall that we output a global shape code $S$ of shape $N \times 1$ in our encoder of the alignment network. We assume that the shape code has encoded the shape information of the input point clouds and, in the ideal case, can directly output the accurate final shape with the decoding operation. Yet as we have mentioned, the information of only one frame is often insufficient to give a complete shape, which inspires us to fuse the shape information at the shape code level first. Therefore, we first utilize a GRU module to fuse the shape and ensure consistency between frames. The GRU memory unit takes the shape-code $S_{final}$ as input, and outputs the updates $S'$ and the hidden state $h$. $S'$ should be the last state of the last layer of $h$. 
	
    The second unit is inserted into the refinement module before GCN layers. Since we do not have a per-point correspondence between $P_s^t$ and $P_s^{t+1}$, we can not employ the GRU for temporal fusion directly on the features $f_{align}$ in this module. Apart from the long-term memory, we also hope to focus more on recent information to refine the converged shape of the first stage in the refinement network. Thus, in the second memory unit, we only utilize the information of the most recent frames and fuse them through a Squeeze-and-Excitation(SE) channel-wise attention module\cite{Hu2020SqueezeandExcitationN}. We set a sliding window of a small size $N$ and restore the aligned point clouds in the time window. At time t, we put $P_{align}^t$ and $P_{align}^{t-i}, i = 1, 2, ..., T$ together into the network $FE$ for feature extraction. We also operate the same network again to extract features from $P_{align}^{t-i}$ within the same radius of points in $P_{in}$. The two features are then concatenated and passed through several stacked linear layers to reduce the number of feature channels and get $f_{align}$. After that, we insert a SE block to fuse the temporal information better. Therefore, the final feature $f_{refine}$ can be formulated as
    \begin{equation}
    \begin{aligned}
        f_{align}^t &= MLP[FE(p, P_{align}^t), FE(p, P_{align}^{t-i})], p \in P_{in} \\
        f_{refine}^t&=SE(f_{align}^t).
    \end{aligned}
    \end{equation}
    
    In our experiment, we select $N = 3$. This attention module is, in fact, not a traditional memory unit in deep learning research. However, together with the sliding window mechanism, it also works as a temporal fusion strategy that enables the point clouds to fuse better locally.
	
    \begin{table*}[ht]
	\centering
	\vspace{1.5em}
	\setlength{\abovecaptionskip}{0pt}%
	\setlength{\belowcaptionskip}{0pt}%
	\caption{Comparison in terms of per-point Chamfer Distance $\times 10^{-4}$(Lower is Better) Upper: Augmented ShapeNet Lower: PCN Test}
	\begin{tabular}{c|c|cccccccc}
		\toprule
		Methods & Average & Watercraft & Plane & Cabinet & Car & Chair & Lamp & Couch & Table\\
		\midrule
		PCN\cite{Yuan2018PCNPC} & 14.28 &13.99 & 7.41 & 17.62 & 11.82 & 16.34 & 14.07 & 17.79 & 15.17\\
		TopNet\cite{Tchapmi2019TopNetSP} & 13.88 &13.34 &7.52 &17.57 &11.98 &16.14 &13.37 &16.89& 14.22\\
		GCC Stage 1\cite{Wen2020PointCC} & 8.05 & 8.62 & 3.42 & 9.58 & 6.42 & 8.84 & 10.55 & 8.64 &8.33\\
		GCC Stage 2 & 9.19 & 9.06 & 4.95 & 10.79 & 7.64 & 9.87 & 11.41 & 9.67 & 9.63\\
		Ours & \bf{6.64} & \bf{7.24} & \bf{2.61} & 8.14 & \bf{5.63} & \bf{7.11} & \bf{8.18} & \bf{7.27} & \bf{6.90}\\
		\midrule
		PCN\cite{Yuan2018PCNPC} & 12.59 &10.96 & 7.48 & 15.85 & 10.65 & 15.43 & 12.37 & 16.02 & 11.96\\
		TopNet\cite{Tchapmi2019TopNetSP} & 12.49 &10.59 &8.04 &16.06 &10.91 &14.85 &11.86 &16.01& 11.61\\
		GCC Stage 1 & 7.85 & 7.46 & 3.88 & 10.34 & 6.48 & 8.43 & 9.67 & 9.09 & 7.43\\
		GCC Stage 2 & 9.24 & 8.86 & 5.57 & 11.54 & 7.88 & 9.50 & 11.38 & 10.64 & 8.52 \\
		Ours & \bf{6.51} & \bf{6.26} & \bf{3.08} & \bf{9.45} & \bf{5.91} & \bf{6.45} & \bf{7.29} & \bf{8.02} & \bf{5.69}\\
		\bottomrule
	\end{tabular}
	\label{shapenet-cd}
	\vspace{-1.5em}
    \end{table*}	
	
	\subsection{Loss Function}
	The training losses can be divided into alignment loss and completion loss. The former is designed for the point cloud alignment task in the first stage, and the latter is almost the same as the losses in GCC\cite{jieqigcc}.
	
	\subsubsection{Alignment Loss}
	 Recall that we output three aligned point clouds and translation matrices in the stage-1 network. During training, we use two different losses for them.
	
	 The first is the earth mover distance(EMD) for point clouds. The EMD loss aims to find a bijection between the reference point cloud P and the target point cloud Q that minimizes the average distance between corresponding points. Considering that there exists randomness during the fps process, directly calculating the distances between point clouds may lead to unexpected errors. Therefore, we use EMD to calculate the distance between $P_{align}$ and corresponding aligned input point clouds $P_{ogt}$ of the same resolution. We take the same realization as in PCN\cite{Yuan2018PCNPC}.

	    \begin{equation}
	        L_{emd} = EMD(P_{align}^0, P_{ogt}^0) + \sum_{i=1}^2 \lambda_i EMD(P_{align}^i, P_{ogt}^i).
	    \end{equation}

	 The other loss is the Huber loss for translation, where we add another term to punish the translation error of aligned point clouds. Here we select the typical smooth-l1(Huber) loss to reduce the effect of outliers:
        \vspace{-0.5em}
	    \begin{equation}
	       L_{Huber} = \sum_{i=0}^3 Huber(norm(T^i - T_{gt}). delta=2.0)
	    \end{equation}

	In the ideal case, $P_{align}^0$ and $P_{ogt}^0$ should be exactly the same, while other pairs may suffer small differences. Thus, we set the weight of the EMD loss for $P_{align}^0$ to 3 times that of other aligned point clouds and expect the network to focus more on them. The translations should be the same for all resolutions. Thus we keep their weight the same.
    
    \subsubsection{Completion Loss}
    Considering that our aligned point clouds may still suffer from transformation, we do not take the matching loss from GCC, but use only two kinds of loss functions instead. Firstly, we use the Chamfer distance(CD) loss. The bi-directional CD loss is used to guide the completion process and to ensure the similarity in shapes between the output and generated shapes.
        \begin{equation}
    	e_{CD}(X, Y) = \frac{1}{N_0}\sum_{x \in X}\min_{y \in Y}||x - y||^2_2 + \frac{1}{N_1}\sum_{y \in Y}\min_{x \in X}||y - x||^2_2.
    	\end{equation}
    	Here we also calculate the CD loss twice for both the stage-1 output $P_r$ and the final output $P_f$.

    Secondly, we use the scale-dependent Laplacian loss\cite{Shen2020InteractiveAO}. According to \cite{jieqigcc}, this loss will guide the network to memorize more details and avoid over-smoothing in a backward manner. It can be formulated as
    	\begin{equation}
	e = \sum_{x \in P_g, y \in A(x)} || x - y||_2,\\
	L_{Lap} = ||\sum_{x \in P_g} \sum_{y \in A(x)} \frac{2(x - y)}{e  ||x - y||_2} ||_2.
	\end{equation}
	
	\subsubsection{Overall Loss}
	The overall loss for one frame is the weighted sum of the above terms,
    
	\begin{equation}
	\begin{aligned}
	L &=(\alpha L_{emd} + L_{huber}) + \beta (e_{CD}(P_r, P_{gt}) \\
	  &+ e_{CD}(P_f, P_{gt})) + \gamma(L_{Lap}).
	\end{aligned}
	\end{equation}
	
	During training, we feed three frames to the network sequentially to train the temporal modules in our framework. Therefore, we also make modifications to the losses accordingly. The main changes are made to the CD loss. Apart from the CD loss between current outputs and the ground-truth point cloud, we calculate the CD loss of subsequent frames to guide the training of temporal modules.
	
	\section{Experiments}
	\subsection{Training Details}
	In the training process, we retain 2048 as the size of the input point cloud for different frames, which is the same as the setting in PCN\cite{Yuan2018PCNPC}. We use two GRU layers to memorize information in the temporal unit of the alignment network. In the stage-2 network, we select 256 as the number of controlling points, which is suggested in \cite{jieqigcc}. 
	
	We train our network from the scratch on the ShapeNet dataset, and apply the best model to experiments on all datasets. We first train the alignment network with batch size 32 for 20 epochs(1.5 hours per epoch), and then fix the parameters to train the refinement network separately with batch 20 on a single frame(2.5 hours per epoch). To refine the temporal module, we input three frames to train the whole pipeline after the training work on a single frame is done for another 12 epochs(3 hours per epoch). All the training is finished using the ADAM optimizer and on one GTX 1080 Ti GPU. In the following sections, all tables without specific explanation are presented with per-point Chamfer Distance $\times 10^{-4}$. 

	\subsection{Datasets}
	\subsubsection{ShapeNet}

    \begin{table*}[htbp]
	\centering
	\vspace{1.5em}
	\setlength{\abovecaptionskip}{0pt}%
	\setlength{\belowcaptionskip}{0pt}%
	\caption{Temporal Completion on Aug. ShapeNet(CD $\times 10^{-4}$).}
	\begin{tabular}{c|ccccccccc}
		\toprule
		First Frame & Second & Third & Forth & Fifth & Sixth & Seventh & Eignth & Ninth & Tenth\\
		\midrule
		6.64 & 5.65 & 5.47 & 5.37 & 5.32 & 5.30 & 5.28 & 5.27 & 5.27 & 5.26\\
		\bottomrule
	\end{tabular}
	\vspace{-2em}
	\label{shapenet-temporal}
    \end{table*}

	\begin{figure}[ht]
		\centering
		
		\framebox{\parbox{3.2in}{
				\centering

				\includegraphics[scale=0.24]{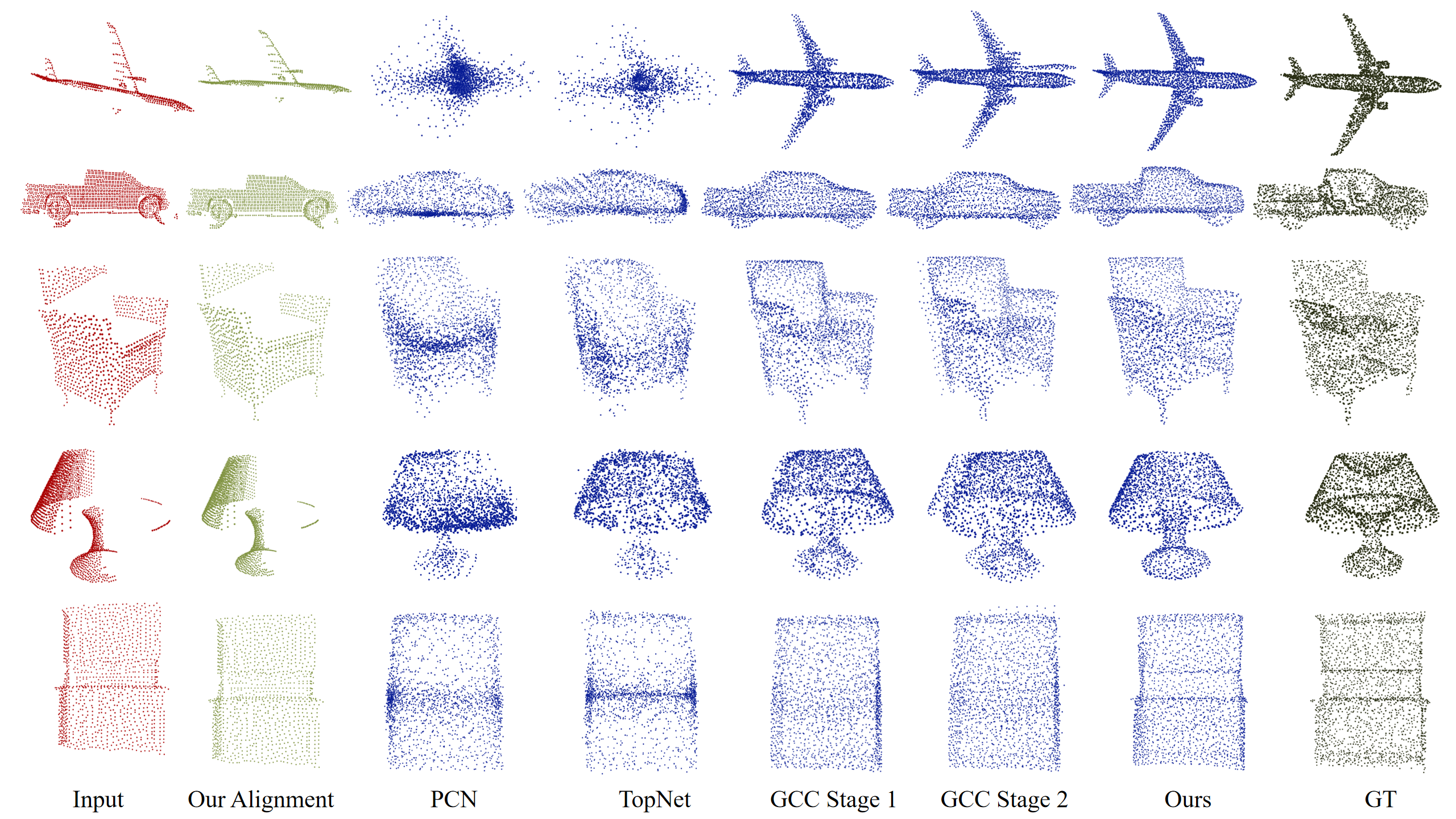}
		}}
		\caption{The single frame completion of the PCN test split. The input point clouds are augmented with random rotation and translation.}
		\label{shapenet}
	\end{figure}

	Following pervious works, we project the CAD models in the ShapeNet dataset to 2D space for partial depth images from different angles, and randomly select 2048 points from the dense depth images as the input. The ground truth data are extracted by uniformly sampling 2048 points on surfaces of the CAD models. There are eight kinds of objects in the ShapeNet dataset, airplanes, cabinets, cars, chairs, lamps, sofas, tables and vessels, and we do not tell the network explicitly what kind of object each input represents. The train, val and test split are the same for PCN\cite{Yuan2018PCNPC}.
	
	To generate the pose disturbance that exists in real-world datasets, we first apply random augmentation to each input point cloud. We rotate the input point cloud in three dimensions, with each rotation angle from -20 to 20, and give the input data a translation from -0.1 to 0.1. We generate 16 depth maps for each CAD model and treat them as a sequence of data collected from the same object. In the training phase, we randomly select three frames from the sequence, apply augmentation to each frame separately, and feed them to the network. In the testing procedure, we operate each test 10 times and get an average result to minimize the bias carried by random augmentation.

    We take the CD as the evaluation metric, which is the same as the metric for SA-Net\cite{Wen2020PointCC} and GCC\cite{jieqigcc}, and we select three benchmarks, the most popular coarse completion method PCN\cite{Yuan2018PCNPC} and TopNet\cite{Tchapmi2019TopNetSP}, and the recent coarse completion method GCC\cite{jieqigcc}. Note that there are many outstanding completion methods that are also open-source, such as GR-Net\cite{Xie2020GRNetGR} and VRE-Net\cite{Pan2021VariationalRP}. However, these focus more on the refinement from 2048 points to 16384-point dense generation, and is not suitable for our task, which stops at the coarse stage. We train the network in our generated dataset with rotation and translation, and use the groundtruth data as the guiding signal. In the comparison stage, we test the four models in both our transformed ShapeNet dataset and the original dataset of PCN. 
    
    From Table \ref{shapenet-cd}, it is clear that our methods perform much better in the face of unaligned data. From the qualitative experiments shown in Figure \ref{shapenet}, we can see that our model performs much better, especially when the rotation and translation are serious. Considering that the GCC\cite{jieqigcc} stage-2 network is sensitive to the input data, we also provide the output of its stage-1 network as a reference.
    
\begin{figure}[ht]
		\centering
		\vspace{-1em}
		\framebox{\parbox{3.2in}{
				\centering

				\includegraphics[scale=0.22]{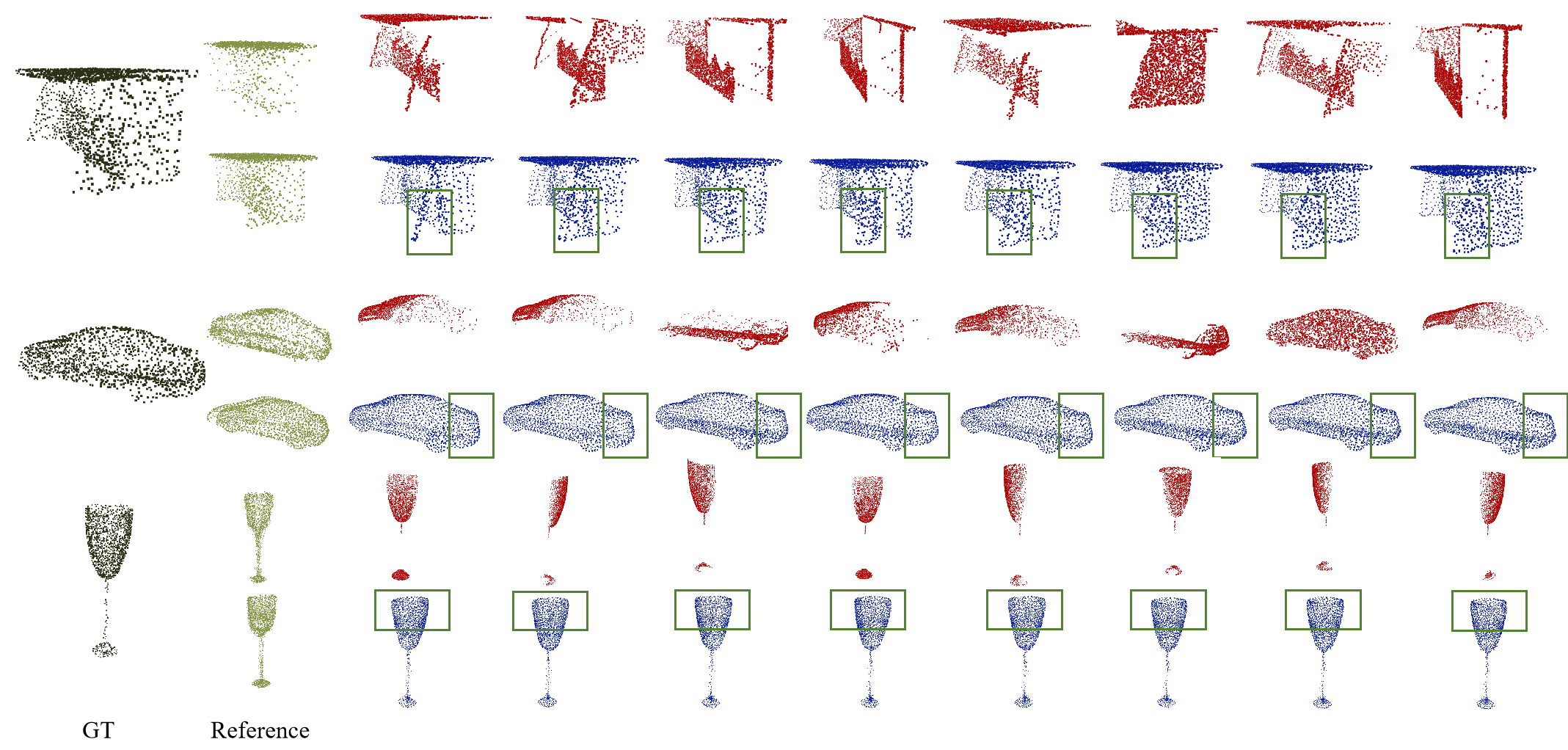}
		}}
		\caption{The temporal completion result in ShapeNet. Red: Input point clouds. Black: GT point clouds. Green: Reference single-frame completion results by TopNet and GCC. Blue: Our temporal completion results. The completion results are arranged from left to right in chronological order.}
		\label{shapenet-qua}
	\end{figure}

    We also show how our temporal module works both quantitatively in Table \ref{shapenet-temporal}, and qualitatively in Figure \ref{shapenet-qua} and \ref{panda}. We can see that the result converges quickly to a satisfactory shape within the first three frames, which indicates that our network can learn complete objects from a small amount of data and make continuous improvement with follow-up information. In the qualitative results, we find that while keeping a consistent shape, the results become better in detailed regions with more input data. 
    \vspace{-1em}
    
    We also evaluate the inference time of the network. For inputs of different batch sizes, the inference time for processing each object is from 43.5 ms(batch size = 1) to 15.3 ms(batch size = 32). We recommend to use batch size = 4 or 8, which can process the data of multiple tracklets at the same time, and ensure that the processing frequency is about 10Hz per batch(20 ms per object). 
    
	\subsubsection{KITTI}
	\begin{figure}[h]
		\centering
		\framebox{\parbox{3.2in}{
				\centering

				\includegraphics[scale=0.24]{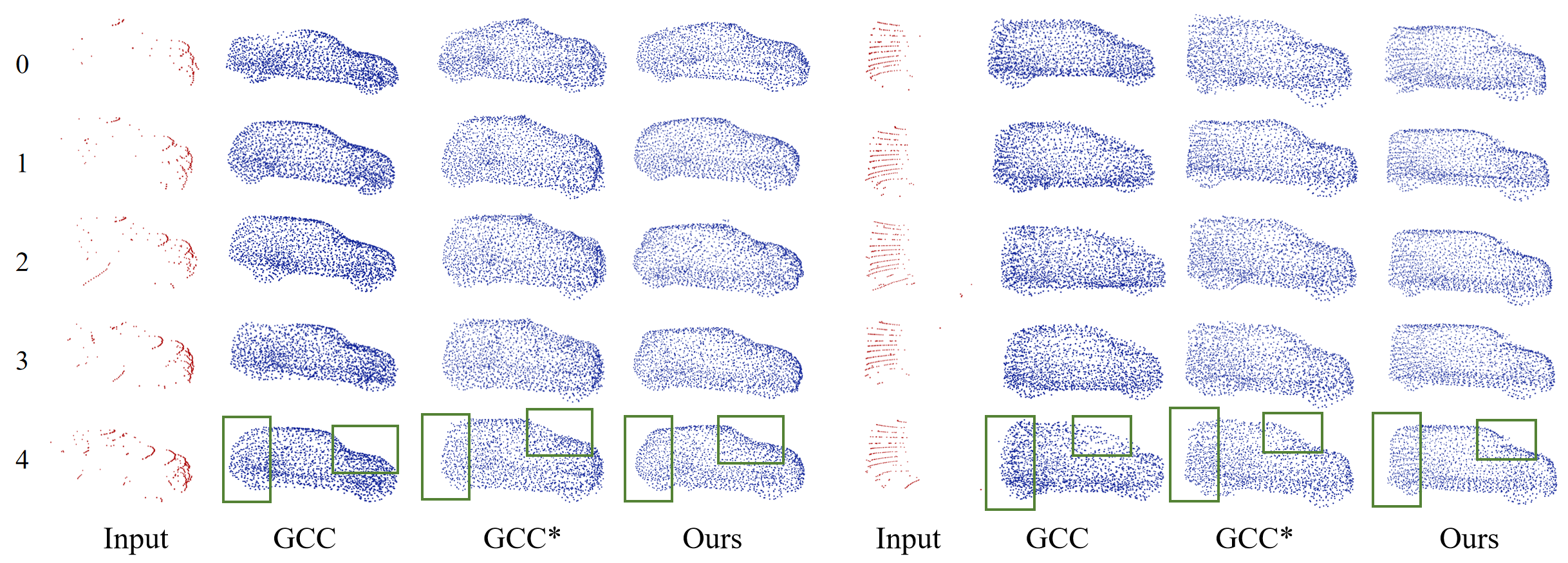}
		}}
		\caption{The temporal completion of image sequences collected from KITTI test split. Poses are provided by the AB3DMOT benchmark. GCC: GCC model with a single input frame. GCC*: GCC model with input point clouds randomly accumulated.}
		\label{kitti}
	\end{figure}

	The KITTI dataset\cite{Geiger2013VisionMR} is one of the most popular self-driving datasets, with real-world scenes collected by lidar sensors. However, there are no grounth truth data for cars or pedestrains in KITTI. In point cloud completion tasks, researchers usually evaluate three key factors in KITTI\cite{Yuan2018PCNPC}fidelity, minimal matching distance(MMD) and consistency. The fidelity measures the average distance between input point clouds and the output, while MMD measures the minimal matching distance between the output result and the ground truth models in the training dataset(the ShapeNet training split). Recalling that we train on the eight categories in the ShapeNet dataset with random pose disturbance, these two indexes, which are designed for aligned input data, are meaningless. We, therefore, only measure the consistency between the output data.
	 
	We first utilize the AB3DMOT\cite{Weng2020_AB3DMOT} tracking model to crop out vehicles and provide an initial 6D pose of the objects. After that, we input the sequence of data into our model and test the consistency between output frames by calculating the CD between two frames. For comparison, we directly accumulate sequential points of each object with provided poses and do random sampling to ensure the size of the input point cloud. We then feed the accumulated point clouds into benchmark models for comparison. We treat five frames as a group and evaluate the consistency between the subsequent frames, obtaining four indexes within a group. In Table \ref{deform-consis}, we show that, except the first and second frame, our method keeps the best consistency and shows an explicit downward trend, which indicates convergence. 
	\vspace{-1em}
	\begin{table}[htbp]
		\setlength{\abovecaptionskip}{0pt}%
		\setlength{\belowcaptionskip}{0pt}%
		\centering
		\caption{Consistency in Kitti Test Split(CD $\times 10^{-4}$).}
		\begin{tabular}{c|ccccccccc}
			\toprule
			Methods & 1-2 & 2-3 & 3-4 & 4-5 \\
			\midrule
			PCN & \bf{4.85} & 5.59 & 4.30 & 3.65\\
		    TopNet & 5.09 & 5.72 & 4.44 & 3.74\\
		    GCC & 5.57 & 6.17 & 4.82 & 4.08 \\
		    Ours & 5.47 & \bf{4.01} & \bf{3.35} & \bf{2.83} \\
			\bottomrule
		\end{tabular}
		\label{deform-consis}
	\end{table}
	\vspace{-1em}
	We also show that, qualitatively, our method obtains a better result in the long run. As is illustrated in Figure \ref{kitti}, the point clouds provided by our model are sampled uniformly and smoothly, and keep more details of the input shape. Due to the disturbance of input information, the long-term model generated by GCC always tends to a square van, while our model can better recover different shapes. We also provide the results obtained with unfused input clouds and denote the different methods as GCC and GCC*. 
	
	\begin{table*}[htbp]
		\setlength{\abovecaptionskip}{0pt}%
		\setlength{\belowcaptionskip}{0pt}%
		\vspace{1em}
		\centering
		\caption{Results with and Without Refinement Layers. Upper: Aug. ShapeNet. Lower: PCN Test(CD $\times 10^{-4}$).}
		\begin{tabular}{c|c|cccccccc}
			\toprule
			Methods & Average & Cabinet & Car & Chair & Lamp & Couch & Table & Watercraft & Plane\\
			\midrule
			W/ Refinement & \bf{6.64} & \bf{7.24} & \bf{2.61} & \bf{8.14} & \bf{5.63} & \bf{7.11} & \bf{8.18} & \bf{7.27} & \bf{6.90}\\
			W/O Refinement & 7.49 & 8.42 & 3.01 & 8.70 & 5.89 & 8.21 & 10.07 & 7.86 & 7.83\\
			\midrule
			W/ Refinement & \bf{6.48} & \bf{6.04} & \bf{3.11} & \bf{9.77} & \bf{5.88} & \bf{6.36} & \bf{7.09} & \bf{7.92} & \bf{5.66}\\
			W/O Refinement & 7.31 & 7.16 & 3.51 & 10.16 & 5.96 & 7.47 & 9.07 & 8.46 & 6.70\\
			\bottomrule
		\end{tabular}
		\vspace{-2em}
		\label{deform-cd}
	\end{table*}

	\section{Ablation Study}
	Apart from the completion task, we perform extended experiments to test the functions of different parts of our network. We also test the quality of the aligned point clouds in the alignment module and how they perform as a registration task. 
	
	\subsection{Efficiency of Refinement Network}
	Similar to \cite{jieqigcc}, we validate the efficiency of our stage-2 refinement network. We first test the per-frame accuracy of our network with and without the refinement module in our augmented ShapeNet test set and the original PCN dataset, and the results are given in Table \ref{deform-cd}. We then show how they perform in the temporal completion task in ShapeNet, with the results shown in Table \ref{deform-temporal}. The alignment module obviously achieves a relatively satifactory result, and illustrates a trend of convergence. Yet the lower bound of the alignment module still does not reach the per-frame completion accuracy of the framework incorporating refinement, which indicates the importance of the refinement network.
    \vspace{-1em}
	\begin{table}[!h]
		\setlength{\abovecaptionskip}{0pt}%
		\setlength{\belowcaptionskip}{0pt}%
		\centering
		\caption{Temporal Completion with and Without Refinement Layers on Aug. ShapeNet(CD $\times 10^{-4}$).}
		\begin{tabular}{c|ccccc}
			\toprule
			Methods & First & Second & Third & Forth & Fifth\\
			\midrule
			W/ Refinement & \bf{6.64} & \bf{5.69} & \bf{5.48} & \bf{5.38} & \bf{5.33}\\
			W/O Refinement & 7.49 & 6.59 & 6.35 & 6.25 & 6.21\\
			\bottomrule
		\end{tabular}
		\label{deform-temporal}
		
	\end{table}
	\vspace{-1em}

	\subsection{Efficiency of the Alignment Module}
	Though we only use a simple T-Net to align the input point clouds, our alignment module seems to provide a satisfactory result that can be used for further refinement. We also evaluate the effect of our alignment network quantitatively in this section.
	
	The first experiment evaluates the distance between our aligned point clouds and the ground-truth input. Since we generate the augmented dataset by applying rotation and translation on the original dataset, we can easily evaluate the alignmenet module on both ShapeNet and PCN datasets. Here, we use both the CD and the EMD between point clouds for evaluation, and also provide the distance between the ground-truth and unaligned point clouds as a reference. Since the alignment module is executed before the temporal fusion module, we use only a single frame for evaluation. All CDs are multiplied by $10^{-4}$. Results are presented in Table \ref{align-cd}.
	\begin{table}[!h]
	    \vspace{-1em}
		\setlength{\abovecaptionskip}{0pt}%
		\setlength{\belowcaptionskip}{0pt}%
		\centering
		\caption{Point Alignment Accuracy in Both Datasets.}
		\begin{tabular}{c|cccc}
			\toprule
			Data Type & PCN CD & PCN EMD & ShapeNet CD & ShapeNet EMD\\
			\midrule
			Aligned& 6.32 & 8.67 & 4.45 & 7.98 \\
			Input& 86.46 & 87.03 & 80.68 & 81.82 \\
			\bottomrule
		\end{tabular}
		\label{align-cd}
	\end{table} 
	
	Another experiment is performed to feed the output of our alignment pont cloud into the benchmark models that are trained on the aligned ShapeNet dataset, and test the results on the PCN test split. We show in Table \ref{align_bench} that though our aligned point clouds can not achieve the same performance as the pre-aligned input point clouds, it outperforms the unaligned input greatly and gives a satisfactory result. 
	\begin{table}[!h]
	\vspace{-1em}
		\setlength{\abovecaptionskip}{0pt}%
		\setlength{\belowcaptionskip}{0pt}%
		\centering
		\caption{Completion on Aligned Points in Aug. PCN Test(CD $\times 10^{-4}$).}
		\begin{tabular}{c|ccc}
			\toprule
			Data Type & PCN\cite{Yuan2018PCNPC} & TopNet\cite{Yuan2018PCNPC} & GCC\\
			\midrule
			W/O Aug. & 10.59 & 9.49 & 6.02\\
			W. Aug. & 12.59 & 12.49 & 7.08\\
			W. Aug. and Orig. Model & 39.79 & 40.37 & 33.42\\
			\bottomrule
		\end{tabular}
		\label{align_bench}
        \vspace{-2em}
	\end{table}

	\subsection{Effect of Input Number}
	In our experiments, we train on 2048 points in the augmented ShapeNet dataset and test on the ShapeNet, PCN and KITTI datasets with different numbers of input point clouds. Though qualitatively we have seen that with at least 20 points in KITTI we can still obtain a good shape, we want to know, quantitatively, how the number of input point clouds will affect the results of our model. Thus, we change the number of input point clouds and test the completion both temporally and with a single frame in ShapeNet and the PCN test dataset.
	
	Our results are listed in Table \ref{num}. We first test the single frame completion result in the augmented PCN test split. We randomly sample N points from the original point cloud, and then stochastically select another 2048 - N points from the N points to lift the input size to 2048. It can be seen that the results do not change significantly with up to 512 points. Even if we decrease the number again to 256, we can still get a satifactory accuracy. Further, we test the temporal accuracy on augmented ShapeNet with five frames. It can be seen that especially with a lack of input information, multi-frame fusion can effectively improve the integrity and accuracy of the final completion. This proves the rationality of our method.
	\begin{table}[!h]
    	\vspace{-1em}
		\setlength{\abovecaptionskip}{0pt}%
		\setlength{\belowcaptionskip}{0pt}%
		\centering
		\caption{Comparison of Input Number(CD $\times 10^{-4}$).}
		\begin{tabular}{c|c|ccccc}
			\toprule
			Input Number & Aug. PCN\cite{Yuan2018PCNPC} & 1 & 2 & 3 & 4 & 5\\
			\midrule
			2048 & 6.48 & 6.64 & 5.65 & 5.47 & 5.37 & 5.32\\
			1024 & 6.60 & 6.91 & 5.89 & 5.62 & 5.49 & 5.44\\
			512 & 6.87 & 7.47 & 5.24 & 5.93 & 5.79 & 5.71\\
			256 & 7.65 & 8.75 & 7.24 & 6.77 & 6.59 & 6.50 \\
			\bottomrule
		\end{tabular}
		\label{num}
		\vspace{-1.5em}
	\end{table}
	
\section{Conclusion}
In this paper, we propose a point cloud completion method that tolerates pose disturbance of limited rotation and translation and employs temporal information for better shape recovery. The noise-tolerance ability of our pipeline enables it to be used in real-world cases where input point clouds may be cropped out by popular detection models that are not as precise as they are in the CAD datasets. Also, by utilizing the temporal information of subsequent frames, our method can output satisfactory results after a short period of initialization using sparse data, and remain consistent between frames while improving the completion quality. Such characteristics make it possible for our model to be integrated into other real-world tasks, such as tracking, visual odometry and scene reconstruction, and to operate in an on-line manner. Experiments on the synthetic ShapeNet and real-world KITTI dataset show the robustness and scalability of our method and its practical value.
\vspace{-1em}
\bibliographystyle{IEEEtran}
\bibliography{egbib}
\end{document}